\documentclass[journal,10pt]{IEEEtran}
\usepackage{amsmath,amssymb,bm,graphicx,booktabs,array,cite}
\usepackage[hidelinks]{hyperref}
\hypersetup{pdftitle={Region--Local Copula Evidence Fusion for Heterogeneous Remote Sensing Change Detection},pdfauthor={JI Zhiyuan, YIN Junjun, YANG Jian},pdfkeywords={copula, heterogeneous change detection, superpixel, evidence fusion}}
\newcommand{\ind}{\mathbb{I}}
\newcommand{\eps}{\varepsilon}

\begin{document}
\title{Region--Local Copula Evidence Fusion for Heterogeneous Remote Sensing Change Detection}
\author{JI Zhiyuan, YIN Junjun, and YANG Jian%
\thanks{JI Zhiyuan and YANG Jian are with the Department of Electronic Engineering, Tsinghua University, Beijing 100084, P.R. China (e-mail: jizy20@mails.tsinghua.edu.cn; yangjian\_ee@tsinghua.edu.cn).}%
\thanks{YIN Junjun is with the School of Computer and Communication Engineering, University of Science and Technology Beijing, P.R. China (e-mail: junjun\_yin@ustb.edu.cn).}}
\markboth{Manuscript Prepared for IEEE Geoscience and Remote Sensing Letters}{Region--Local Copula Evidence Fusion}
\maketitle
\begin{abstract}
Superpixel copula models provide stable regional evidence for heterogeneous remote sensing change detection, but a single label per region limits localization within mixed superpixels. This letter develops a region--local copula evidence fusion method that retains the regional decision structure of SCOPS while introducing spatially varying local dependence anomalies. Independently fitted local models characterize departures from unchanged cross-image relationships. Reference ranking and an upper-tail gate transform these anomalies for fusion with continuous regional confidence. We derive the resulting region-dependent local decision threshold and identify a condition under which gating is equivalent to reparameterizing ungated fusion. On Lake and UK, whole-image optimized configurations achieve kappa coefficients of 0.78136 and 0.90817 and improve mixed-region and boundary decisions. Four-fold retrospective spatial validation over ten training subsets confirms complementary local information, with ungated reference fusion increasing mean kappa by 0.00693 and 0.01793. Fixed gating yields a larger UK gain of 0.03353 but only 0.00041 on Lake. These results support regional--local dependence interaction, while showing that calibration and gating have scene-dependent benefits.
\end{abstract}

\section{Introduction}
\IEEEPARstart{H}{eterogeneous} remote sensing images acquired with different spectral responses can differ radiometrically even over unchanged surfaces. Consequently, local intensity discrepancies need not indicate land-cover change. Copula models separate marginal distributions from cross-image dependence and offer a statistical way to characterize unchanged relationships. Conditional copula change detection already combines dependence modeling with local statistical comparisons \cite{mercier}; structural approaches, such as iterative robust graphs, exploit relationships among image elements instead of directly comparing intensities \cite{irg}.

SCOPS provides a superpixel-based copula construction with explicit family selection and prior-dependent statistics \cite{scops}. Regional aggregation reduces local fluctuations and supports compact modeling, but its final labels are constant within each superpixel. A region intersecting a change boundary may contain both classes, which no single regional label can represent. Increasing regional confidence precision alone does not remove this spatial restriction: the score still has one value throughout the region.

Recent copula methods address other aspects of heterogeneous detection. COMIC combines copula mixtures with cycle-consistent image translation \cite{comic}; NN-Copula-CD learns dependence using a copula-constrained neural network \cite{nncopula}; and FAR-Aware-Copula-CD designs a copula-based test for false-alarm control \cite{far}. Here, the objective is to retain an established regional detector while enabling decisions that vary within its spatial support. Local copula statistics, rather than direct radiometric differences, provide the required additional evidence without an image-translation network.

We propose region--local copula evidence fusion, denoted A2 in the experiments. Its contribution has two connected aspects. First, independently fitted regional and local dependence models combine regional stability with within-region discrimination. Second, an explicit decision analysis shows how regional confidence sets the amount of local anomaly evidence required to declare change, and clarifies the role and limits of reference ranking and gating. Whole-image comparisons, controlled evidence ablations, and retrospective spatial validation assess these mechanisms separately. Copula estimation, empirical ranking, and convex fusion are established operations; the contribution concerns their spatial support and interaction in extending SCOPS.

\section{Region--Local Copula Evidence Fusion}
\subsection{Two Spatial Supports and a Shared Statistical Construction}
Let $X\in\mathbb R^{H\times W\times C_X}$ and $Y\in\mathbb R^{H\times W\times C_Y}$ be registered, normalized images. Their joint superpixel segmentation gives regions $S_i$, with $s(p)$ the region containing pixel $p$. Regional and local features are
\begin{align}
 x^{\mathrm{sp}}_{i,a}&=\frac{1}{|S_i|}\sum_{p\in S_i}X_a(p),&
 y^{\mathrm{sp}}_{i,b}&=\frac{1}{|S_i|}\sum_{p\in S_i}Y_b(p),\label{eq:regionalmeans}\\
 x^{\mathrm{loc}}_a(p)&=\frac{1}{w^2}\sum_{t\in\Omega_w(p)}X_a(t),&
 y^{\mathrm{loc}}_b(p)&=\frac{1}{w^2}\sum_{t\in\Omega_w(p)}Y_b(t).
 \label{eq:localmeans}
\end{align}
Here $w$ is odd and border pixels are replicated. Regional features are constant on $S_i$, whereas overlapping local windows move with $p$. The two branches use the same selected unchanged training regions, but their marginal distributions and copulas are fitted \emph{separately}. Training-patch and full-image local means are filtered separately with the same padding rule, as in the implementation.

For a channel pair $(a,b)$ and either scale $r\in\{\mathrm{sp},\mathrm{loc}\}$, the unchanged joint density satisfies
\begin{equation}
 f_{XY,0}^{r}(x,y)=c_0^{r}(F_{X,0}^{r}(x),F_{Y,0}^{r}(y))
 f_{X,0}^{r}(x)f_{Y,0}^{r}(y),\label{eq:sklar}
\end{equation}
with channel indices suppressed. Thus the copula density describes cross-image dependence after marginal transformation. Low density indicates incompatibility with the learned unchanged dependence; its negative logarithm is an anomaly statistic, potentially negative, rather than a probability.

For $n$ unchanged samples, fitting uses $u_j=\operatorname{rank}(x_j)/n$ and $v_j=\operatorname{rank}(y_j)/n$, with ties assigned by the implementation's sorting order. Evaluation retains SCOPS' insertion-rank rule,
\begin{equation}
 \widehat F^{\mathrm{ins}}_{X,0}(x)=
 \frac{1+\sum_{j=1}^{n}\ind[x_j\le x]}{n+2},\quad
 u=\widehat F^{\mathrm{ins}}_{X,0}(x),\label{eq:margin}
\end{equation}
and analogously for $v$. Negative training Kendall correlation is handled by $v\leftarrow1-v$ in fitting and evaluation; endpoint values are clipped into $(0,1)$. For each pair and scale, SCOPS selects among $\mathcal Q=\{\mathrm{Gaussian},t,\mathrm{Clayton},\mathrm{Gumbel}\}$:
\begin{equation}
 k_{ab}^{r,*}=\arg\max_{k\in\mathcal Q}\frac{1}{n}
 \sum_{j=1}^{n}\log c_k(u_j,v_j;\widehat\theta_{ab,k}^{r}),
 \label{eq:selection}
\end{equation}
where $\widehat\theta$ is fitted using \texttt{copulafit}. Write the selected density as $\widehat c_{ab}^{r}$. The inherited pair statistic is
\begin{equation}
 T_{ab}^{r}(x,y)=-\log\widehat c_{ab}^{r}(u,v)
 -\delta\log\widehat f_{Y,b,0}^{r}(y).\label{eq:stat}
\end{equation}
Prior~1 uses $\delta=0$; Prior~3 uses $\delta=1$ and a three-component Gaussian mixture for $\widehat f_Y$. For the unreflected transform, dividing \eqref{eq:sklar} by $f_{X,0}$ gives
\begin{equation}
 f_{Y|X,0}(y|x)=c_0(F_{X,0}(x),F_{Y,0}(y))f_{Y,0}(y),
 \label{eq:conditional}
\end{equation}
which motivates the additional term in Prior~3. The implementation keeps this statistic after the negative-correlation reflection. Both densities are floored at a positive machine constant before taking logarithms. The family selection in \eqref{eq:selection} is distinct from the following maximum across channel-pair anomalies.

\subsection{Continuous Regional Evidence}
SCOPS aggregates the pair statistics and clusters regional features:
\begin{align}
 D_i&=\max_{a,b}T_{ab}^{\mathrm{sp}}(x^{\mathrm{sp}}_{i,a},y^{\mathrm{sp}}_{i,b}),\label{eq:di}\\
 \bm f_i&=[(\bm x_i^{\mathrm{sp}})^\top,(\bm y_i^{\mathrm{sp}})^\top,
                   \alpha(D_i-\bar D)]^\top,\label{eq:feature}
\end{align}
where $\bar D$ is the mean over test regions. Two-class K-means yields assignments $g_i$ and centroids $\bm\mu_k$. The original area-weighted changed-cluster identification is retained:
\begin{equation}
 k_c=\arg\max_{k\in\{1,2\}}\sum_i|S_i|D_i\ind[g_i=k],
 \quad z_i=\ind[g_i=k_c].\label{eq:cluster}
\end{equation}
The other cluster is $k_u$. Whereas the original decision is $z_{s(p)}$, A2 uses a continuous score. With $d_{i,k}=\|\bm f_i-\bm\mu_k\|_2^2$, the distance confidence is
\begin{equation}
 p_i^{\mathrm{sp}}=\frac{d_{i,k_u}}{d_{i,k_c}+d_{i,k_u}+\eps},
 \qquad P_{\mathrm{sp}}(p)=p_{s(p)}^{\mathrm{sp}}.\label{eq:distance}
\end{equation}
A region closer to the changed centroid receives higher confidence. This mapping is used for UK and for both scenes in spatial validation.

The optimized Lake configuration instead uses signed difference-image (DI) confidence. Let $m_k=\operatorname{median}_{z_i=k}D_i$, $\tau_D=(m_0+m_1)/2$, and $s_D=|m_1-m_0|/2+\eps$. Then
\begin{equation}
 p_i^{\mathrm{sp}}=\sigma\left((2z_i-1)\frac{|D_i-\tau_D|}{s_D}\right),
 \qquad \sigma(t)=(1+e^{-t})^{-1}.\label{eq:signed}
\end{equation}
The cluster label determines the sign and the DI distance determines the magnitude. Neither mapping estimates a calibrated posterior. Crucially, both retain a score constant within each superpixel; continuous regional confidence provides a fusion interface but cannot by itself resolve a mixed region.

\subsection{Local Dependence Evidence and Reference Ranking}
Local model fitting uses centers sampled from the selected unchanged training superpixels. Applying \eqref{eq:stat} to \eqref{eq:localmeans} gives
\begin{equation}
 T_{\mathrm{loc}}(p)=\max_{a,b}T_{ab}^{\mathrm{loc}}
                 (x^{\mathrm{loc}}_a(p),y^{\mathrm{loc}}_b(p)).\label{eq:local}
\end{equation}
Evaluate the same fitted models at the training reference centers, using \eqref{eq:margin} again, and aggregate identically to obtain $T_{0,1},\ldots,T_{0,N_0}$. This ensures that reference and test anomalies share the same evaluation rule. Regional parameters are not substituted into the local model.

We rank each local anomaly relative to the unchanged references:
\begin{equation}
 P_{\mathrm{loc}}(p)=\frac{1}{N_0+1}\sum_{j=1}^{N_0}
                    \ind[T_{0,j}\le T_{\mathrm{loc}}(p)].\label{eq:cdf}
\end{equation}
This increasing empirical percentile lies in $[0,N_0/(N_0+1)]$. It measures how many unchanged references have weaker anomalies, not the survival probability $1-F$ or a posterior change probability. For any common strictly increasing $h$, $T_{0,j}\le T$ if and only if $h(T_{0,j})\le h(T)$; hence the score is invariant to monotone rescaling of the statistic. This removes arbitrary amplitude scaling, but does not guarantee improved class separation.

An upper-tail gate suppresses weak local responses:
\begin{equation}
 A_q(p)=\left[\frac{P_{\mathrm{loc}}(p)-q}{1-q}\right]_+,
 \quad [t]_+=\max(0,t),\quad 0\le q<1.\label{eq:gate}
\end{equation}
For $P_{\mathrm{loc}}\le q$, local evidence is zero; above $q$, its magnitude increases linearly. Ungated reference fusion is recovered at $q=0$. Reference samples are reused for fitting and ranking, and neighboring windows are dependent. Accordingly, $q$ is an evidence-admission quantile, not a guaranteed significance level or false-alarm rate.

\begin{figure}[t]
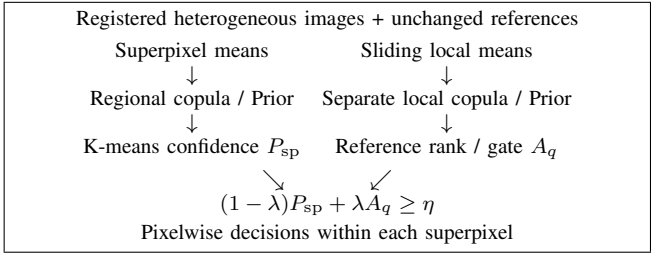
\centering\footnotesize
\fbox{\parbox{.94\columnwidth}{\centering
Registered heterogeneous images + unchanged references\\[4pt]
\begin{tabular}{c@{\quad}c}
Superpixel means & Sliding local means\\
$\downarrow$ & $\downarrow$\\
Regional copula / Prior & Separate local copula / Prior\\
$\downarrow$ & $\downarrow$\\
K-means confidence $P_{\mathrm{sp}}$ & Reference rank / gate $A_q$
\end{tabular}\\[4pt]
$\searrow\qquad\qquad\swarrow$\\
$(1-\lambda)P_{\mathrm{sp}}+\lambda A_q\ge\eta$\\[2pt]
Pixelwise decisions within each superpixel}}
\caption{A2 framework. The regional branch preserves context, while the independently fitted local branch allows evidence to vary within a region. Reference anomaly ranking is applied only to the local branch.}
\label{fig:framework}
\end{figure}

\subsection{Fusion and the Induced Within-Region Decision Rule}
The final evidence score and binary decision are
\begin{align}
 R(p)&=(1-\lambda)P_{\mathrm{sp}}(p)+\lambda A_q(p),\label{eq:fusion}\\
 \widehat C(p)&=\ind[R(p)\ge\eta],\quad 0\le\lambda\le1.\label{eq:decision}
\end{align}
Fig.~\ref{fig:framework} summarizes the construction. At $\lambda=0$, A2 reduces to thresholded continuous regional confidence, which need not equal the original hard SCOPS labels at an arbitrary $\eta$. At $\lambda=1$, only local evidence remains. For $p,p'\in S_i$,
\begin{equation}
 R(p)-R(p')=\lambda[A_q(p)-A_q(p')],\label{eq:variation}
\end{equation}
so the local branch is exactly the source of within-region score variation. Such variation permits, but does not guarantee, different binary labels: the scores must also straddle $\eta$.

For $\lambda>0$, rearranging \eqref{eq:fusion}--\eqref{eq:decision} yields the region-dependent local threshold
\begin{equation}
 \widehat C(p)=1\ \Longleftrightarrow\ A_q(p)\ge h_i,
 \qquad h_i=\frac{\eta-(1-\lambda)p_i^{\mathrm{sp}}}{\lambda}.
 \label{eq:threshold}
\end{equation}
Writing $P_{\max}=N_0/(N_0+1)$ and $A_{\max}=[(P_{\max}-q)/(1-q)]_+$ makes the three cases explicit:
\begin{equation}
 \widehat C(p)=
 \begin{cases}
 1, &h_i\le0,\\
 \ind[P_{\mathrm{loc}}(p)\ge q+(1-q)h_i],&0<h_i\le A_{\max},\\
 0,&h_i>A_{\max}.
 \end{cases}\label{eq:cases}
\end{equation}
In the middle case, stronger regional confidence lowers the required local percentile. For $0<\lambda<1$, its derivative with respect to $p_i^{\mathrm{sp}}$ is $-(1-q)(1-\lambda)/\lambda<0$. Thus regional context changes the local evidence requirement instead of enforcing one label for all pixels. This is an algebraic property of the score rule, not a Bayesian posterior derivation.

\subsection{When Gating Changes Only the Parameterization}
Gating need not define a distinct binary decision family. A sufficient condition is $\eta>1-\lambda$: for $P_{\mathrm{loc}}\le q$, $R\le1-\lambda<\eta$, so no changed decision is possible. For $P_{\mathrm{loc}}>q$, multiplying the decision inequality by $1-q$ gives
\begin{equation}
 (1-q)(1-\lambda)P_{\mathrm{sp}}+\lambda P_{\mathrm{loc}}
 \ge (1-q)\eta+\lambda q.\label{eq:expand}
\end{equation}
Define $B=(1-q)(1-\lambda)+\lambda>0$ and
\begin{equation}
 \lambda'=\frac{\lambda}{B},\qquad
 \eta'=\frac{(1-q)\eta+\lambda q}{B}.\label{eq:equiv}
\end{equation}
Dividing \eqref{eq:expand} by $B$ gives ungated fusion $(1-\lambda')P_{\mathrm{sp}}+\lambda'P_{\mathrm{loc}}\ge\eta'$. Below $q$, its left side is also strictly below $\eta'$, since \eqref{eq:expand}'s left side is at most $(1-q)(1-\lambda)+\lambda q$. The decisions are therefore identical everywhere. Equal-size finite grids in $(\lambda,\eta)$ need not cover equal decision boundaries after this transformation. We consequently evaluate ranking and gating as conditional design choices, while attributing the spatial mechanism to regional--local evidence interaction.

\section{Experiments and Results}
\subsection{Data, Settings, and Evaluation Protocols}
We use the Lake and UK benchmarks described in SCOPS \cite{scops}. Lake covers expansion in Sardinia at $300\times412$ pixels and 30-m resolution. Its inputs are a Landsat-5 near-infrared image (September 1995) and a three-channel optical image supplied through Google Earth (July 1996), as dated in the benchmark description. UK covers flooding in Gloucester at $990\times554$ pixels and 25-m resolution, with a three-channel SPOT image (1999) and a post-event NDVI image (2000). The supplied registration and reference maps are retained. Normalization and joint SLIC segmentation follow SCOPS. Ground truth (GT) selects pure unchanged training superpixels under the inherited SCOPS reference-selection protocol; no changed-class model is trained.

The training strips comprise columns 1--130 on Lake and 420--469 on UK (one-based). Retaining 70\% of eligible unchanged regions gives 704/1005 and 216/308 training superpixels. Local reference samples are capped at 4000 and 1500. Both final runs use $w=3$, segmentation seed~1, and 20 K-means++ replicates. Lake uses Prior~1, $\alpha=3$, 2000 requested test superpixels, and SLIC compactness~1; UK uses Prior~3, $\alpha=5$, 2600 superpixels, and compactness~0.4. We report kappa coefficient (KC), F1 score, and overall accuracy (ACC), with change as the positive class.

\emph{Whole-image reference protocol:} The established A2 results use $(q,\lambda,\eta)=(.97,.225,.55)$ for Lake and $(.96,.65,.64)$ for UK, with signed-DI and distance confidence, respectively. Parameters and training-subset seed~1 were selected by whole-image GT metrics. These are optimized reference results, distinct from held-out validation. The public comparison contains only the original published SCOPS row; other baselines are also quoted through \cite{scops}, rather than rerun under a common protocol.

\emph{Retrospective spatial protocol:} Each image is divided into four geometric quadrants. Entire superpixels are excluded if they cross quadrant boundaries, intersect a two-pixel band around the dividing lines, or intersect the training strip expanded by two pixels. The remaining evaluation areas contain 74234 Lake and 441457 UK pixels, with no superpixel shared between folds. Both scenes use distance confidence and $3\times3$ local means. In each fold, three quadrants select fusion parameters and the fourth evaluates them. Selection maximizes mean validation KC over all ten training-subset seeds, with mean F1 and ACC as tie breakers; no seed is selected. The four held-out confusion matrices are pooled separately for each seed, then metrics are summarized by mean and sample standard deviation over seeds.

All fused variants search $\lambda\in\{0,.1,\ldots,1\}$ and $\eta\in\{0,.025,\ldots,1\}$, totaling 451 settings; region-only uses the same threshold grid. The primary gate is fixed at $q=.97$, with $.95$ and $.99$ evaluated separately. These scenes and backbone settings participated in earlier development, and full-image unlabeled features remain available. This protocol therefore evaluates spatial transfer of fusion parameters, not generalization to an unseen scene. GT-assisted reference selection and GT-assisted parameter selection are reported separately.

\begin{figure*}[t]\centering
\includegraphics[width=.98\textwidth]{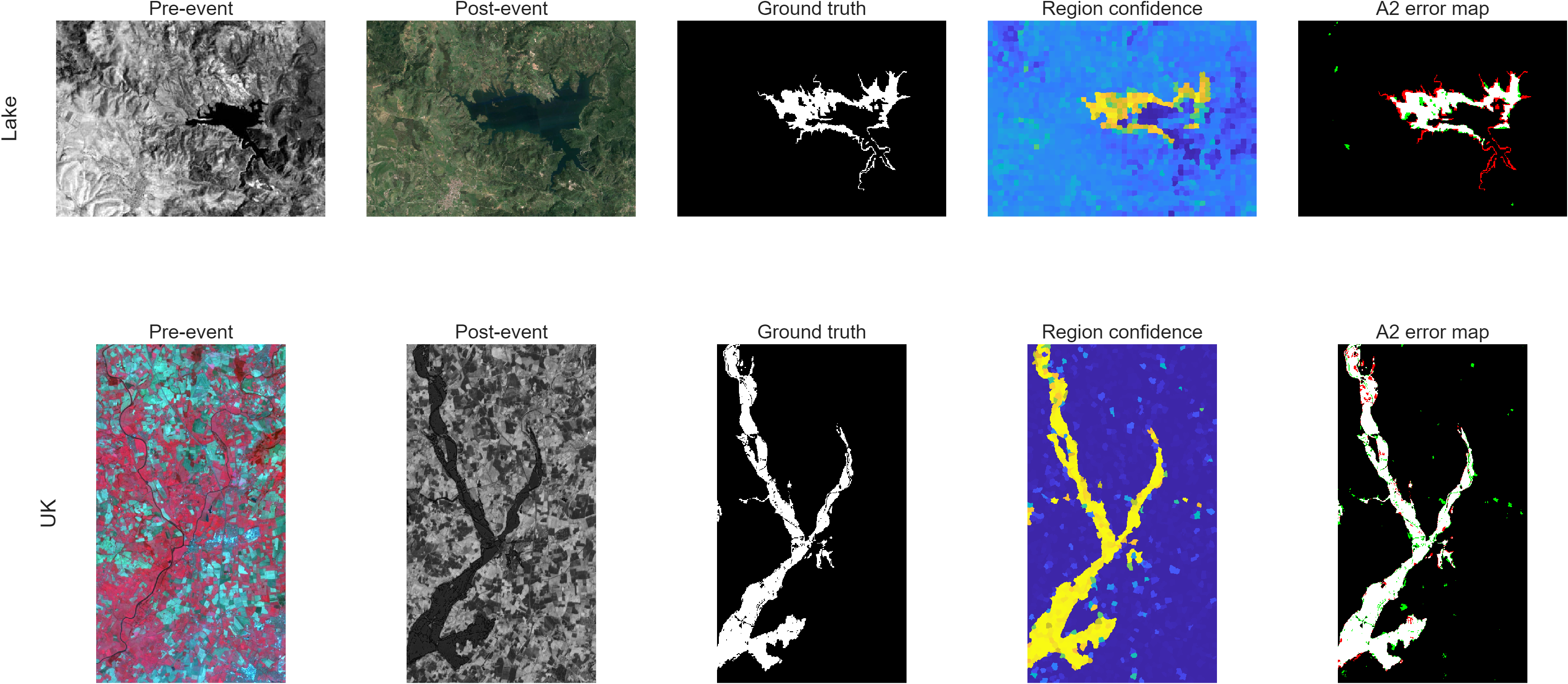}
\caption{Whole-image optimized A2 results on Lake (top) and UK (bottom). Columns show the image pair, GT, regional confidence (cool to warm: low to high), and A2 error map. White, black, green, and red denote true positives, true negatives, false positives, and false negatives, respectively. These are the established reference runs, not out-of-fold predictions.}
\label{fig:qualitative}
\end{figure*}

\begin{table}[t]\centering\footnotesize
\caption{Whole-image comparison. Baselines are published values quoted through SCOPS; A2 is the optimized reference run.}
\label{tab:main}\setlength{\tabcolsep}{2.8pt}
\begin{tabular}{lrrrrrr}\toprule
&\multicolumn{3}{c}{Lake}&\multicolumn{3}{c}{UK}\\
\cmidrule(lr){2-4}\cmidrule(lr){5-7}
Method&KC&F1&ACC&KC&F1&ACC\\\midrule
M3CD&.669&.689&.963&.588&.636&.915\\
FPMS&.552&.588&.925&.816&.837&.962\\
NPSG&.559&.587&.947&.608&.663&.902\\
IRG-McS.dist&.739&.754&.971&.714&.749&.939\\
IRG-McS.sim&.733&.749&.971&.735&.768&.942\\
SCOPS&.764&.778&.973&.859&.876&.970\\\midrule
\textbf{A2}&\textbf{.78136}&\textbf{.79361}&\textbf{.97667}&\textbf{.90817}&\textbf{.91937}&\textbf{.98032}\\\bottomrule
\end{tabular}
\end{table}

\subsection{Whole-Image Results and Evidence Complementarity}
Table~\ref{tab:main} reports the established reference results. A2 exceeds the quoted SCOPS KC by .01736 on Lake and .04917 on UK; differences in evaluation and parameter-selection protocols preclude treating these as matched estimates of generalization gain. With fixed A2 parameters over ten training subsets, KC is $.77372\pm.00449$ and $.89747\pm.00766$. Twenty external K-means seeds reproduce the final seed-1 metrics to numerical precision.

Internal component ablations yield region-only/local-only KC of .76287/.75628 on Lake and .87815/.89119 on UK. Complete A2 exceeds both branches, supporting complementarity. Each component was optimized within its existing grid, rather than switched off at an unchanged threshold. In a separate controlled experiment with min--max regional DI, $w=3$, common reference ranking, and the same gating/fusion search, copula evidence reaches KC .78095/.90454 (Lake/UK). The best pixel or local-mean radiometric difference reaches only .76359/.87700, with optimal local weight .05 in both scenes. This distinguishes useful local dependence information from merely adding spatially varying intensity differences.

\subsection{Mixed-Region and Boundary Localization}
Fig.~\ref{fig:qualitative} shows the regional confidence and final errors. For a matched internal analysis, we compare A2 with SCOPS run on the same training samples and random settings. A mixed superpixel contains both GT classes; the boundary band is within two pixels of the GT change boundary. Table~\ref{tab:spatial} shows that A2 increases accuracy on both spatial subsets in both scenes. Net corrections equal corrected baseline errors minus newly introduced errors. Gains concentrate on UK, consistent with its larger local fusion weight.

\begin{table}[t]\centering\footnotesize
\caption{Internal paired localization analysis. S denotes matched SCOPS, used here only for spatial mechanism assessment.}
\label{tab:spatial}\setlength{\tabcolsep}{3.2pt}
\begin{tabular}{llrrrr}\toprule
Scene&Subset&Pixels&S ACC&A2 ACC&Net gain\\\midrule
Lake&Mixed&14357&.78422&.80685&325\\
Lake&Boundary&9195&.69897&.71452&143\\
UK&Mixed&89811&.84197&.89535&4794\\
UK&Boundary&40416&.73013&.81173&3298\\\bottomrule
\end{tabular}
\end{table}

For small changed connected components with area no larger than the median test-superpixel area, UK gains 120 correct pixels among 1230. However, neither method detects the 70 pixels in Lake's four small components. The evidence supports improved mixed-region and boundary localization, not universal recovery of small objects or increased physical image resolution.

\subsection{Spatial Validation and Calibration Analysis}
Table~\ref{tab:blocked} presents out-of-fold KC. Ungated reference-CDF fusion improves over region-only in all ten seeds on both scenes, with mean gains .00693 and .01793. Fixed gating improves UK by .03353 in all ten seeds, but its Lake mean gain is only .00041, with six seeds improving and four declining. The large optimized Lake gain therefore does not persist under the common regional mapping and fixed gate. Neither pixel nor local-mean differences improve mean KC, reinforcing the importance of dependence evidence.

\begin{table}[t]\centering\footnotesize
\caption{Retrospective blocked validation: out-of-fold KC mean $\pm$ sample standard deviation over ten training subsets. All variants use regional distance confidence.}
\label{tab:blocked}\setlength{\tabcolsep}{3pt}
\begin{tabular}{lcc}\toprule
Local evidence / mapping&Lake&UK\\\midrule
Region only&$.76853\pm.00675$&$.85911\pm.00672$\\
Copula / min--max&$.77363\pm.00374$&$.89366\pm.00762$\\
Copula / reference CDF&$.77546\pm.00735$&$.87704\pm.00811$\\
Copula / CDF + gate .97&$.76895\pm.00250$&$.89265\pm.00752$\\
Pixel diff. / CDF + gate .97&$.76832\pm.00712$&$.85702\pm.00710$\\
Mean diff. / CDF + gate .97&$.76786\pm.00702$&$.85677\pm.00710$\\\bottomrule
\end{tabular}
\end{table}

Reference calibration and gating do not consistently beat conventional min--max scaling of $T_{\mathrm{loc}}$. Ungated CDF performs best among these mappings on Lake, whereas min--max slightly exceeds fixed gating on UK. Table~\ref{tab:calibration} separately reports whole-image calibration ablations under the established regional mappings, selecting parameters by ten-subset mean KC. Gating improves over ungated CDF in that grid, but remains below min--max on Lake. Applying reference-CDF calibration to the \emph{regional} statistic also reduced fused KC, from .78095 to .75708 on Lake and .90454 to .88774 on UK in the earlier controlled study; it is therefore not adopted in the regional branch.

\begin{table}[t]\centering\footnotesize
\caption{Whole-image calibration ablation with established regional mappings and a common full threshold grid. Parameters maximize ten-subset mean KC (oracle selection).}
\label{tab:calibration}
\begin{tabular}{lcc}\toprule
Local mapping&Lake KC&UK KC\\\midrule
Min--max&$.77649\pm.00384$&$.89617\pm.00661$\\
Reference CDF&$.76755\pm.00361$&$.88870\pm.00641$\\
Reference CDF + gate .97&$.77313\pm.00423$&$.89769\pm.00717$\\\bottomrule
\end{tabular}
\end{table}

The equivalence in \eqref{eq:equiv} is realized by the optimized UK parameters: $\lambda'=.97891566$ and $\eta'=.97831325$ reproduce its gated binary map at all 548460 pixels. Lake does not satisfy the sufficient condition; the same transformation changes 242 labels. Thus differences on regular parameter grids partly reflect their coverage of decision boundaries, not an intrinsically new classifier supplied by the gate. In spatial validation, $q=.95,.97,.99$ gives Lake KC .76250, .76895, .77151 and UK KC .88979, .89265, .87090; increasing tail selectivity is not uniformly beneficial.

\subsection{Computation and Scope}
The stored final MATLAB runs take 2.386~s on Lake and 7.421~s on UK, including local-model stages of .960~s and 5.218~s. Additional time over the measured SCOPS backbone is 1.159~s and 5.655~s. These are recorded implementation timings, not hardware-normalized benchmarks. Stage-boundary memory increments are 2.00 and 26.32~MB, rather than peak memory. Reconstructed local statistics reproduced stored reference scores and final maps exactly across all ten subsets per scene.

Only two developed image pairs are available. Seed variability measures sensitivity to reference sampling, not uncertainty across independent scenes. The comparison does not establish superiority over recent neural or false-alarm-controlled copula methods, for which matched runs are unavailable. A decisive next evaluation is to fix regional mapping, local calibration, and parameter-selection rules before accessing a new scene, then compare recent copula methods under the same reference and evaluation protocol.

\section{Conclusion}
Regional confidence and local dependence anomalies provide complementary information for heterogeneous change detection. Their interaction produces a region-dependent local decision threshold and enables finer decisions within mixed superpixels. Existing experiments support improved boundary localization and useful spatial transfer, particularly on UK. The derivation and calibration ablations also show why ranking and gating should not be credited with universal gains: gating can merely reparameterize a decision boundary, and its preferred setting depends on the scene. Future validation should test fixed design choices on new image pairs and determine when regional confidence is sufficient versus when local evidence is necessary.

\section*{Acknowledgment}
OpenAI Codex assisted with drafting the abstract and Sections I--IV, organizing mathematical explanations from the existing method and code, and preparing the LaTeX manuscript. Reported measurements were taken from the project's existing experimental records.

\end{document}